\documentclass[10pt,twocolumn,letterpaper]{article}
\usepackage{cvpr}
\usepackage{times}
\usepackage{epsfig}
\usepackage{newclude}
\usepackage{graphicx}
\usepackage{amsmath}
\usepackage{amssymb}
\usepackage[pagebackref=true,breaklinks=true,letterpaper=true,colorlinks,bookmarks=false]{hyperref}
\usepackage{multirow}
\usepackage{graphicx}
\usepackage{subfig}
\usepackage{array}

\cvprfinalcopy 
\def\cvprPaperID{2253} 

\begin{document}
\title{``Zero-Shot'' Super-Resolution using Deep Internal Learning}
\author{Assaf Shocher$^*$
\qquad
Nadav Cohen$^\dagger$
\qquad
Michal Irani$^*$
}

\date{
    $^*$Dept. of Computer Science and Applied Math,
The Weizmann Institute of Science, Israel\\
    $^\dagger$School of Mathematics,
Institute for Advanced Study,
Princeton, New Jersey\\
 \\  
Project Website:   \href{http://www.wisdom.weizmann.ac.il/~vision/zssr/}{http://www.wisdom.weizmann.ac.il/$\sim$vision/zssr/}
}

\maketitle

\begin{abstract}
\label{sec:abstract}
Deep Learning has led to a dramatic leap in Super-Resolution (SR) performance in the past few years. However, being supervised, these SR methods are restricted to specific training data, where the acquisition of the low-resolution (LR) images from their high-resolution (HR) counterparts is predetermined (e.g., bicubic downscaling), without any distracting artifacts (e.g., sensor noise, image compression, non-ideal PSF, etc). Real LR images, however, rarely obey these restrictions, resulting in poor SR results by SotA (State of the Art) methods. In this paper we introduce ``Zero-Shot'' SR, which exploits the power of Deep Learning, but does not rely on prior training. We exploit the internal recurrence of information inside a single image, and \textbf{train a small image-specific CNN at test time, on examples extracted solely from the input image itself}. As such, it can adapt itself to different settings per image. This allows to perform SR of real old photos, noisy images, biological data, and other images where the acquisition process is unknown or non-ideal. On such images, our method outperforms SotA  CNN-based SR methods, as well as previous unsupervised SR methods.
To the best of our knowledge, this is the first unsupervised CNN-based SR method.
%
%
\end{abstract}

\begin{figure*}
\centering
\begin{tabular}{cc}
\multicolumn{2}{c}{\textbf{(a) Historic image: Check-point Charlie (end of World-War II) -- $SR\times2$} }  \vspace*{-0.15cm}\\
\multirow{2}{0.75\textwidth}{\subfloat{\includegraphics[width=0.75\textwidth]{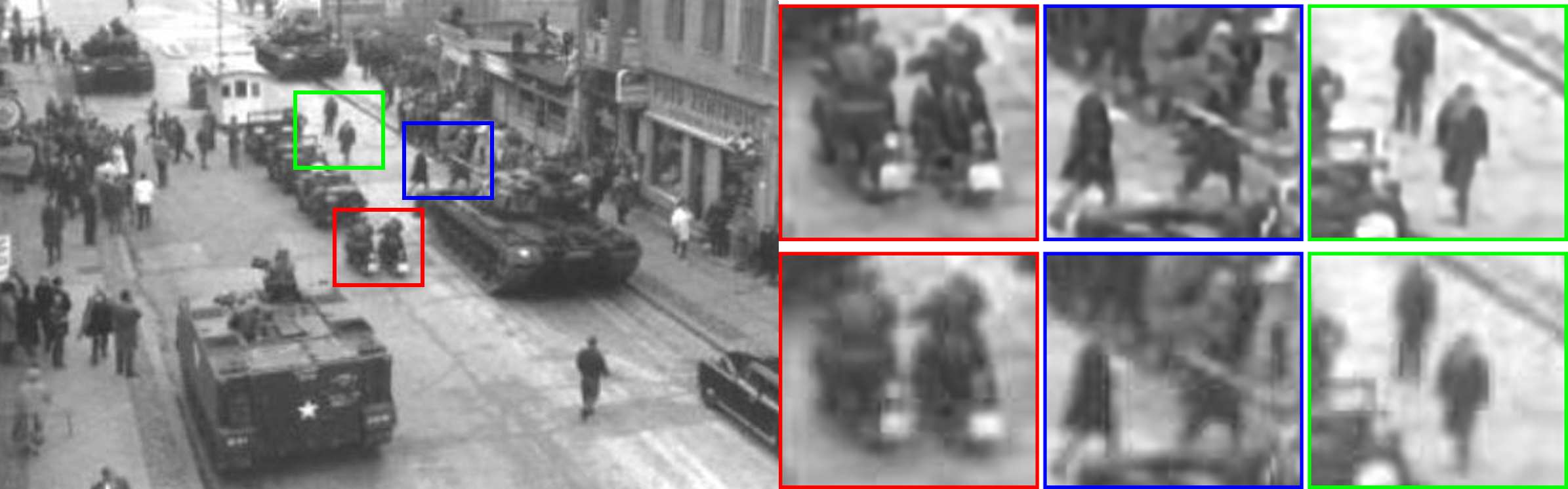}}} \\
 \vspace{0.1\textwidth}
 & ZSSR (ours) \\
 & EDSR \cite{EDSR} \\
  \vspace{0.1\textwidth}
\vspace*{-0.3cm}
\end{tabular}
\begin{tabular}{cc}
\textbf{(b) iPhone image -- $SR\times3$} & \textbf{(c) Historic image: JFK funeral -- $SR\times2$}
\vspace*{-0.5cm}\\
\multirow{5}{0.285\textwidth}{
\subfloat{\includegraphics[width=0.285\textwidth]{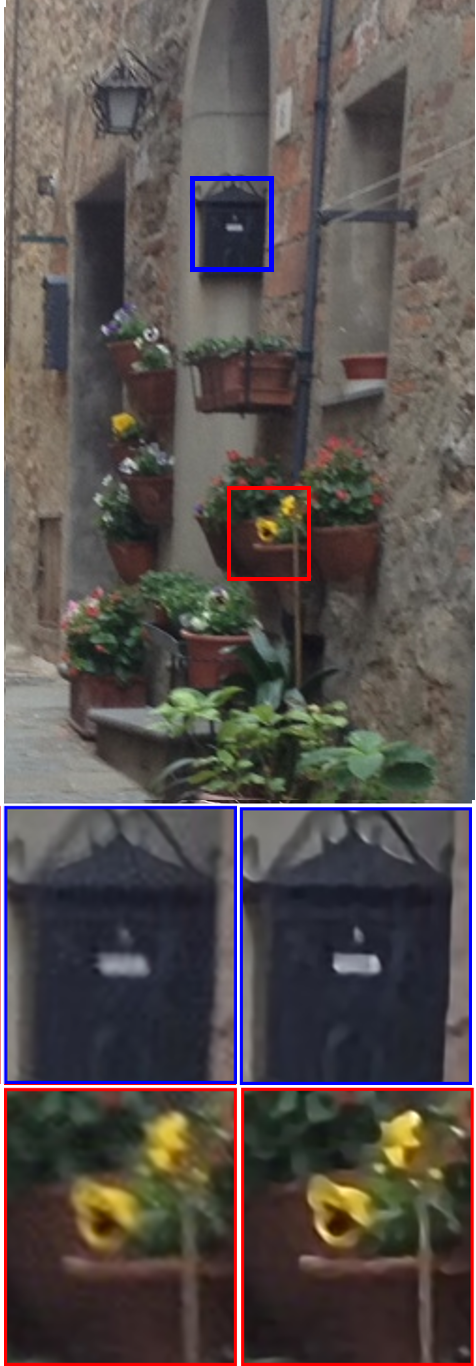}}} \\
  & \subfloat{\includegraphics[width=0.538\textwidth]{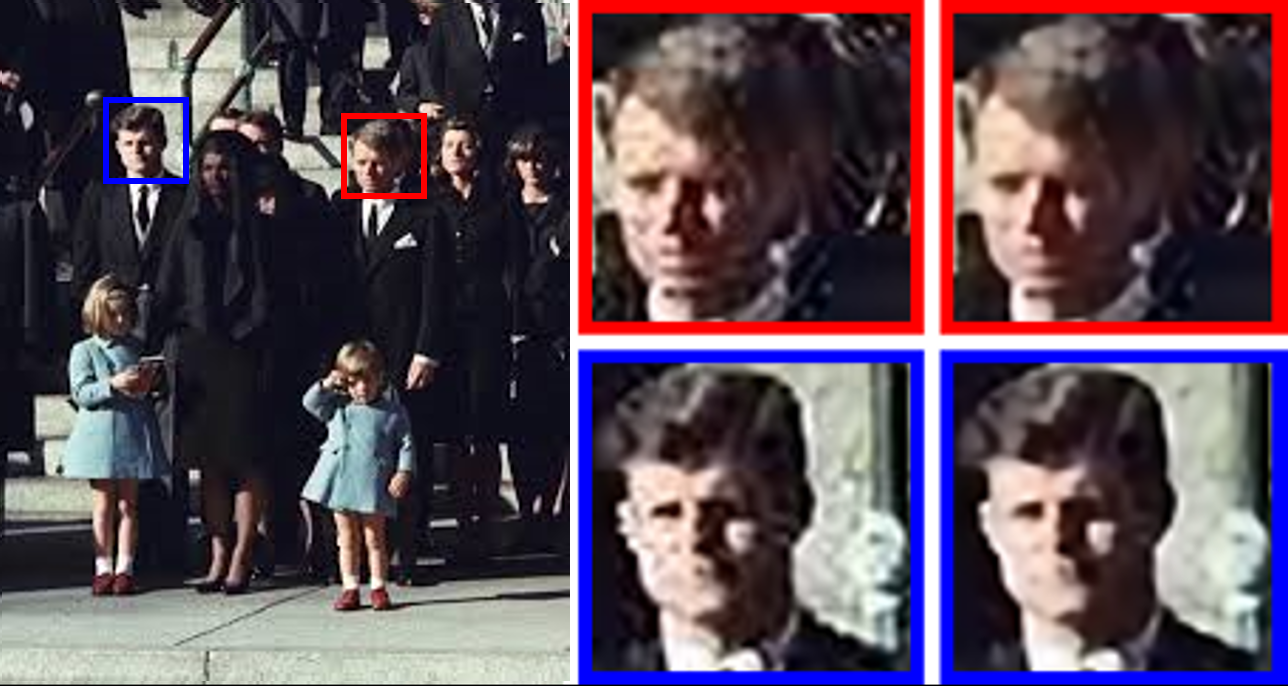}} \\
  & \hspace{0.25\textwidth} EDSR \cite{EDSR}  \hspace{0.05\textwidth} ZSSR (ours) \\
  & \vspace*{-0.15cm} \\
  & \textbf{(d) Outdoor image downloaded from the Internet --  $SR\times2$} \vspace*{-0.15cm}\\
  & \subfloat{\includegraphics[width=0.538\textwidth]{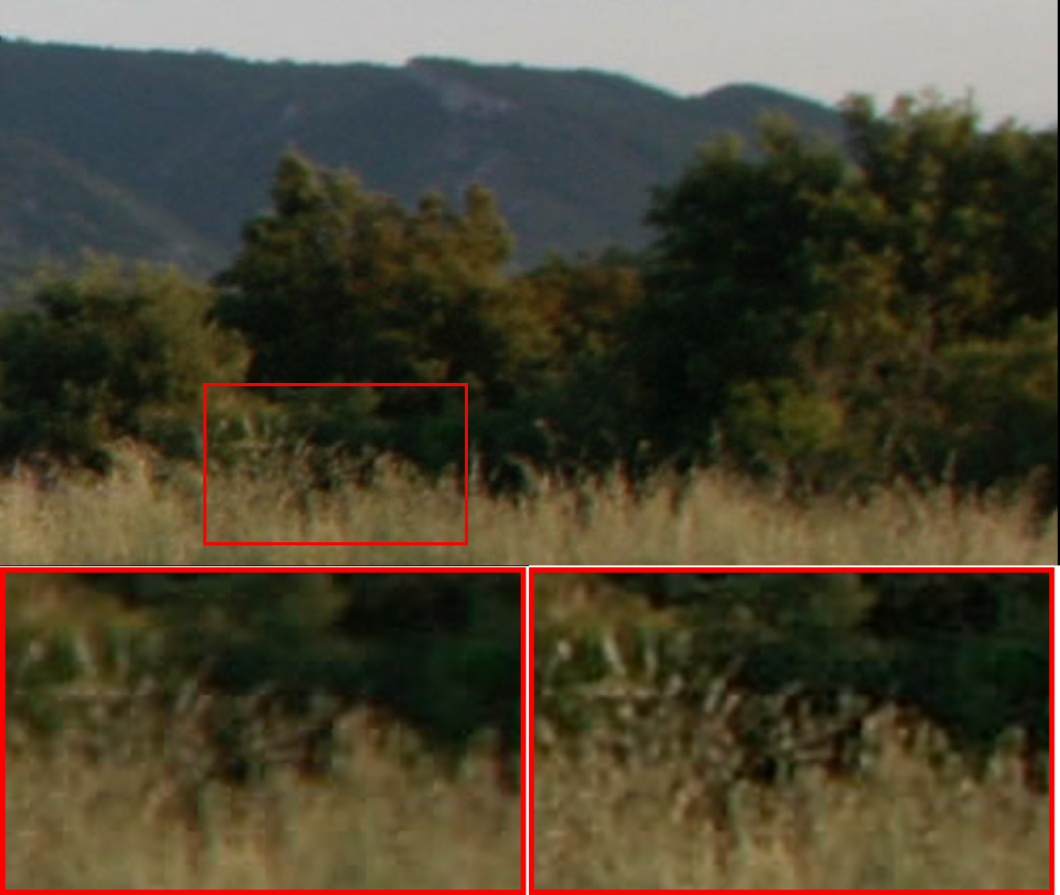}} \\
  EDSR \cite{EDSR}  \hspace{0.05\textwidth}  ZSSR (ours) & EDSR \cite{EDSR}  \hspace{0.2\textwidth}  ZSSR (ours)
\end{tabular}
\caption{\label{fig:figure1}
\textbf{SR of real images (unknown LR acquisition process).}
\emph{Real-world images rarely obey the `ideal conditions' assumed by supervised SR methods. For example, old historic photos~(a,c), images taken by smartphones~(b), random images on the Internet~(d), etc.
Since ZSSR trains at test time on examples extracted from the test image,
it is better at performing SR `In-the-Wild' (i.e., in unconstrained and unknown settings). Full sized images can be found on our \href{http://www.wisdom.weizmann.ac.il/~vision/zssr/}{project website}.
}}
\end{figure*}

 \begin{figure*}
\hspace*{1.8cm}
\textbf{(a)~SR under aliasing:}
  \begin{center}
  \includegraphics[width=14cm]{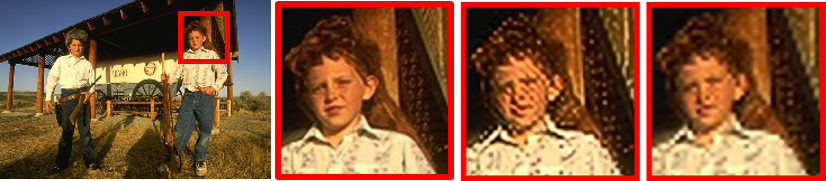}\\
\begin{tabular}{{p{4.4cm}p{2.8cm}p{2.8cm}p{2.8cm}}}
   &\centering Ground truth & \centering EDSR+ \cite{EDSR} &  \centering ZSSR (ours)\tabularnewline
   &\centering (PSNR /SSIM) & \centering ( 21.64 / 0.6641) & \centering (25.02 / 0.7658)\tabularnewline
\end{tabular}\\
\end{center}
\hspace*{1.8cm}
\textbf{(b)~SR under unknown \emph{non-ideal} downscaling kernel:}
\begin{center}
  \includegraphics[width=14cm]{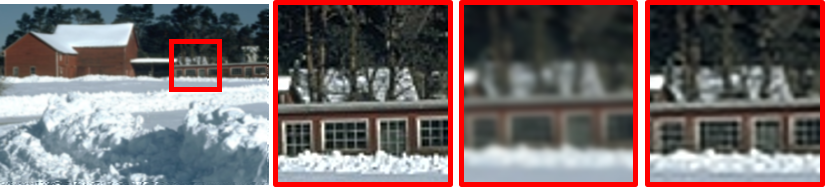}\\
\begin{tabular}{{p{4.4cm}p{2.8cm}p{2.8cm}p{2.8cm}}}
   &\centering Ground truth & \centering EDSR+ \cite{EDSR} &  \centering ZSSR (ours)\tabularnewline
   &\centering (PSNR /SSIM) &\centering (24.44 / 0.7006) & \centering (27.62 / 0.8367)\tabularnewline
\end{tabular}
\vspace*{-0.5cm}
\end{center}
\caption{\label{fig:figure2}
\textbf{SR of `non-ideal' LR images -- a controlled experiment.} \emph{(a)~LR image generated with aliasing (donwscaling kernel is a delta function). (b)~LR image generated with a \emph{non-ideal} downscaling kernel. The \underline{unknown} image-specific kernel is estimated \underline{directly from the LR test image} using~\cite{Michaeli:ICCV13}, and fed into our image-specific CNN as the downscaling kernel (note that externally-trained networks cannot make use of such image-specific information at test-time).
Full sized images can be found on our \href{http://www.wisdom.weizmann.ac.il/~vision/zssr/}{project website}. Quantitative evaluation on hundreds of `non-ideal' LR images can be found in Sec. \ref{sec:experiments_results_nonideal}.
}}

\end{figure*}

\section{Introduction}
\label{sec:introduction}
Super-Resolution (SR) from a single image has recently received a huge boost in performance using Deep-Learning based methods~\cite{SRCNN,DRCN,VDSR,SRGAN,EDSR}. The recent SotA (State of the Art) method~\cite{EDSR} exceeds previous \emph{non}-Deep SR methods (supervised~\cite{A+} or unsupervised~\cite{Glasner,Fattal,SelfEx}) by a few dBs -- a huge margin! This boost in performance was obtained with very deep and well engineered CNNs, which were trained exhaustively on external databases, for lengthy periods of time (days or weeks). However, while these externally  supervised\footnote{We use the term \textbf{\emph{``supervised''}}
for any method that trains on \textbf{\emph{externally supplied examples}} (even if their generation 
does not require manual labelling).}
methods perform extremely well on data satisfying the conditions they were trained on, their performance deteriorates significantly once these conditions are not satisfied.

For example, SR CNNs are typically trained on high-quality natural images,
from which the low-resolution (LR) images were generated
with a specific predefined downscaling kernel (usually a Bicubic kernel with antialiasing -- MATLAB's default imresize command), without any distracting artifacts (sensor noise, non-ideal PSF, image compression, etc.), and for a predefined SR scaling-factor (usually $\times2$, $\times3$ or $\times4$; assumed equal in both dimensions).
Fig.~\ref{fig:figure2} shows what happens when these conditions are not satisfied, e.g., when the LR image is generated with a \emph{non-ideal} (non-bicubic) downscaling kernel, or contains aliasing effects, or simply contains sensor noise or compression artifacts.
Fig.~\ref{fig:figure1} further shows that these are not contrived cases, but rather occur often when dealing with \emph{real} LR images --  images downloaded from the internet, images taken by an iPhone, old historic images, etc. In those `non-ideal' cases, SotA SR methods often produce poor results.

In this paper we introduce ``Zero-Shot'' SR (ZSSR), which exploits the power of Deep Learning, without relying on any prior image examples or prior training. We exploit the internal recurrence of information within a single image and train a small \emph{image-specific} CNN at test time, \emph{on examples extracted solely from the LR input image itself} (i.e., internal self-supervision). As such, the CNN can be adapted to different settings per image. \emph{This allows to perform SR on real images where the acquisition process is unknown and non-ideal}
(see example results in Figs.~\ref{fig:figure1} and~\ref{fig:figure2}). On `non-ideal' images, our method outperforms externally-trained SotA SR methods by a large margin.

The recurrence of small pieces of information (e.g., small image patches) \emph{across scales} of a single image, was shown to be a very strong property of natural images~\cite{Glasner,Zontak:CVPR11}. This formed the basis for many \emph{unsupervised} image enhancement methods, including unsupervised SR~\cite{Glasner,Fattal,SelfEx}, Blind-SR~\cite{Michaeli:ICCV13} (when the downscaling kernel is unknown), Blind-Deblurring~\cite{Michaeli:ECCV14,Bahat:ICCV17}, Blind-Dehazing~\cite{Bahat:ICCP16}, and more.
While such unsupervised methods can exploit image-specific information (hence are less subject to the above-mentioned supervised restrictions), they typically rely on simple Eucledian similarity of small image patches, of predefined size (typically $5 \times 5$), using K-nearest-neighbours search. As such, they do not generalize well to patches that do not exist in the LR image, nor to new implicitly learned similarity measures, nor can they adapt to non-uniform sizes of repeating structures inside the image.

Our image-specific CNN leverages on the power of the \emph{cross-scale} internal recurrence of image-specific information,
without being restricted by the above-mentioned limitations of patch-based methods.
We train a CNN to infer complex image-specific HR-LR relations from the LR image and its downscaled versions (self-supervision). We then apply those learned relations on the LR input image to produce the HR output.
This outperforms unsupervised patch-based SR by a large margin.

Since the visual entropy inside a single image is much smaller than in a general external collection of images~\cite{Zontak:CVPR11}, a small and simple CNN suffices for this image-specific task. Hence, even though our network is trained at test time, its \emph{train+test runtime} is comparable to the \emph{test runtime} of SotA supervised CNNs.  Interestingly, our image-specific CNN produces impressive results (although not SotA) on the `ideal' benchmark datasets used by the SotA supervised methods (even though our CNN is small and has not been pretrained), and \emph{surpasses SotA supervised SR by a large margin
on `non-ideal' images}. We provide both visual and empirical evidence of these statements.

The term ``Zero-Shot'' used here, is borrowed from the domains of recognition/classification. Note however, that
unlike these approaches for \emph{Zero-Shot Learning}~\cite{zero-shot:CVPR2017} or \emph{One-shot Learning}~\cite{one-shot-learning:ICML2016}, our approach does not require any side information/attributes or any additional images.  We may only have a single test image at hand, one of a kind, and nothing else. Nevertheless, when additional information is available and provided (e.g., the downscaling kernel can be estimated directly from the test image using~\cite{Michaeli:ICCV13}), our image-specific CNN can make good use of this at test time, to further improve the results.

\vspace{0.78cm}
\noindent
\underline{Our contributions are therefore several-fold:} \\
(i)~To our best knowledge, this is the first \emph{unsupervised} CNN-based SR method.\\
(ii)~It can handle non-ideal imaging conditions, and a wide variety of images and data types (even if encountered for the first time). \\
(iii)~It does not require pretraining and can be run with modest amounts of computational resources. \\
(iv)~It can be applied for SR to any size and theoretically also with any aspect-ratio. \\
(v)~It can be adapted to known as well as unknown imaging conditions (at test time). \\
(v)~It provides SotA SR results on images taken with `non-ideal' conditions, and competitive results on `ideal' conditions for which SotA supervised methods were trained on. 


\section{The Power of Internal Image Statistics}
\label{sec:image_specific_information}

Fundamental to our approach is the fact that natural images have strong internal data repetition. For example, small image patches (e.g., $5$$\times$$5$, $7$$\times$$7$) were shown to repeat many times inside a single image, both within the same scale, as well as across different image scales.
This observation was empirically verified by~\cite{Glasner,Zontak:CVPR11} using hundreds of natural images,
 and was shown to be true for almost any small patch in almost any natural image.
%

\begin{figure}
  \centering
  \hspace*{-0.2cm} \includegraphics[width=1.1\columnwidth]{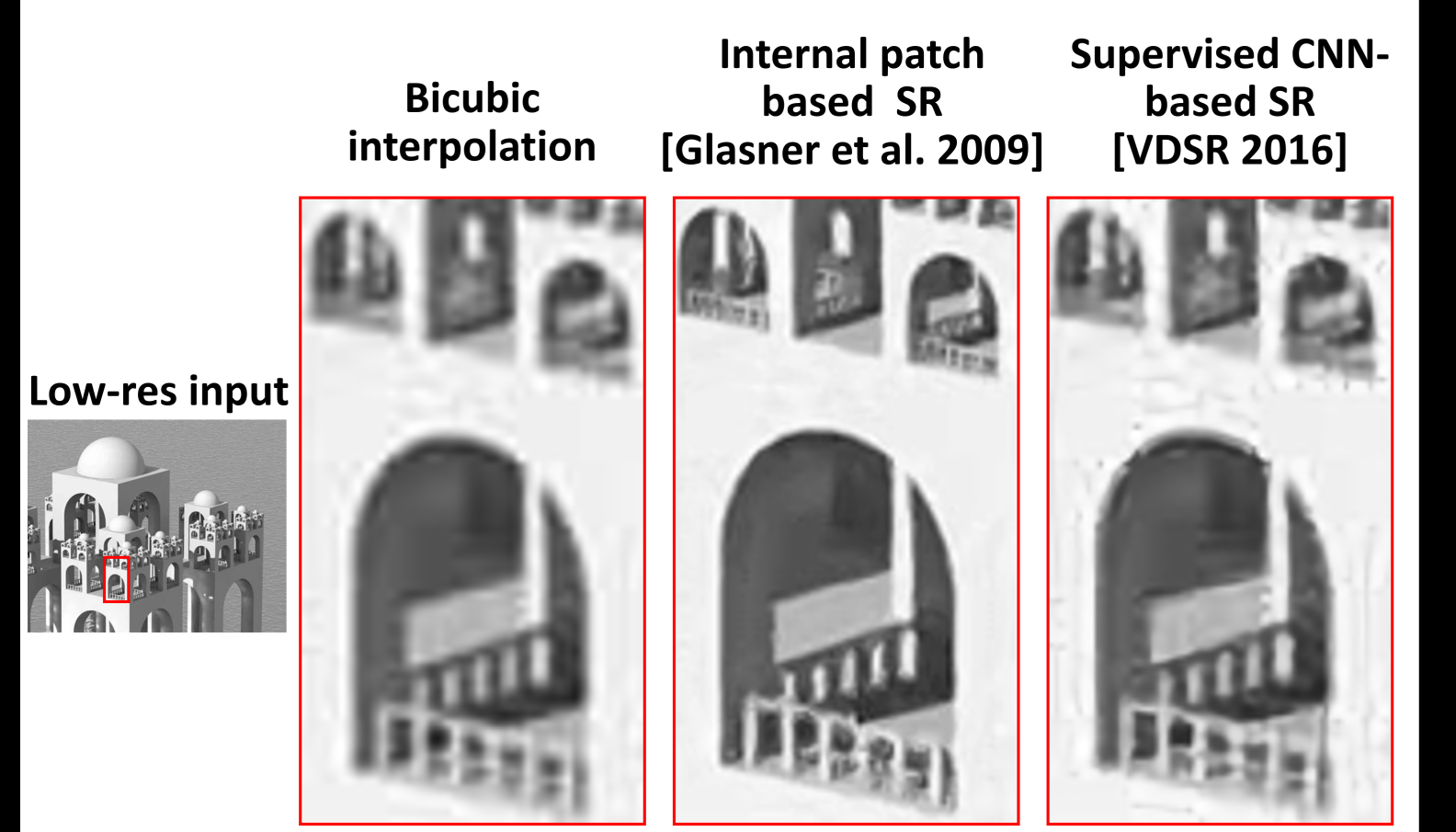}
\caption{\label{fig:invsout}
\textbf{Internal predictive power of image-specific information.}
\emph{Simple unsupervised internal-SR~\cite{Glasner} is able to reconstruct the tiny handrail in the tiny balconies, whereas externally-trained SotA SR methods fail to do so. Evidence to the existence of those tiny handrails exists only internally, inside this image, at a different location and scale (in one of the larger balconies). Such evidence is not found in any external database of images (no matter how large it is).}
}
\end{figure}

Fig.~\ref{fig:invsout} shows an example of a simple single-image SR based on internal patch recurrence (courtesy of~\cite{Glasner}). Note that it is able to recover the tiny handrails in the tiny balconies, since evidence to their existence is found elsewhere inside this image,
in one of the larger balconies.
In fact, \emph{the only evidence to the existence of these tiny handrails exists internally, inside this image, at a {different location} and {different scale}.
It cannot be found in any external database of examples, no matter how large this dataset is!}
As can be seen, SotA SR methods fail to recover this \emph{image-specific information} when relying on externally trained images.
While the strong internal predictive-power is exemplified here using a `fractal-like' image, the internal predictive-power was analyzed and shown to be strong for almost any natural image~\cite{Glasner}.

In fact, it was empirically shown by~\cite{Zontak:CVPR11} that the \emph{internal entropy} of patches inside a single image is much smaller than the \emph{external entropy} of patches in a general collection of natural images. This further gave rise to the observation that internal image statistics often provides \emph{stronger predictive-power} than external statistics obtained from a general image collection. This preference was further shown to be \emph{particularly strong under growing uncertainty and image degradations} (see~\cite{Zontak:CVPR11,Mosseri:ICCP13} for  details).

\begin{figure*}
\centering
  \includegraphics[width=15cm]{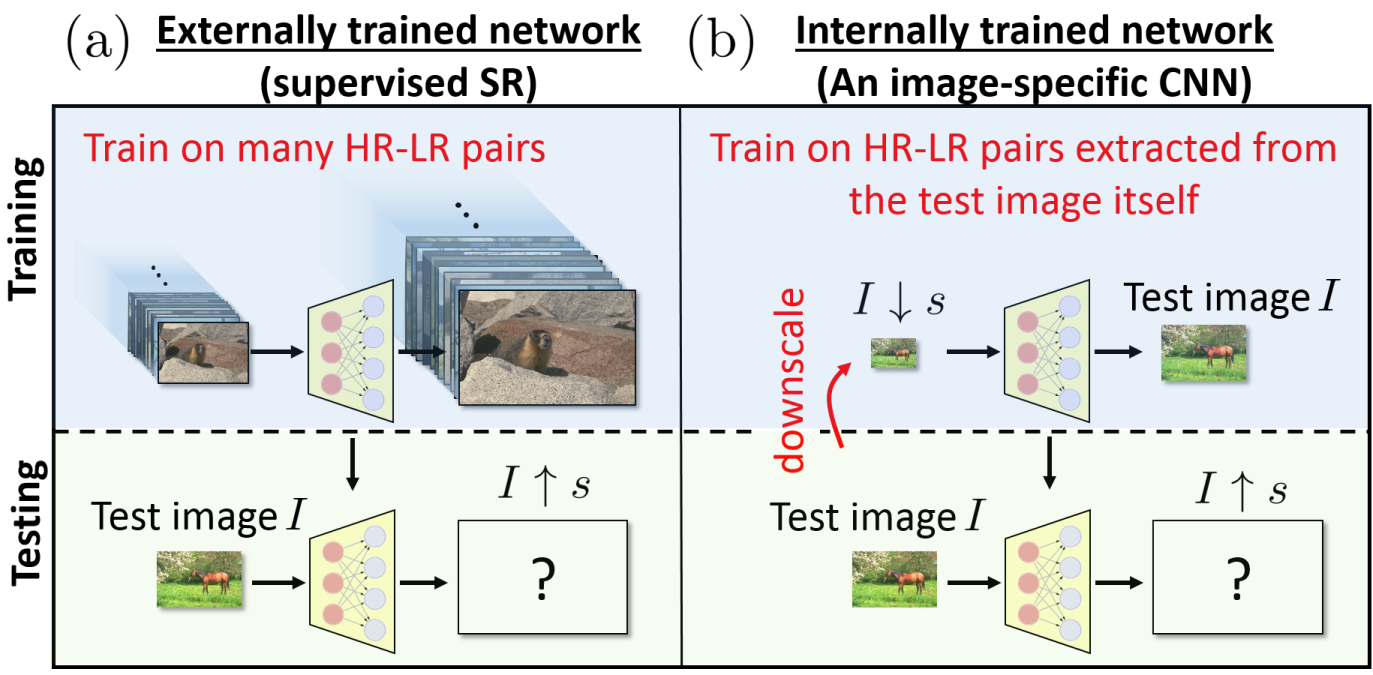}
  \caption{\label{fig:imspecific}
  \textbf{Image-Specific CNN -- ``Zero-Shot'' SR.} 
  \ \emph{(a)~Externally-supervised SR CNNs are pre-trained on large external databases of images. The resulting very deep network is then applied to the test image $I$. \ (b)~Our proposed method (ZSSR): a small image-specific CNN is trained on examples extracted internally, from the test image itself. It learns how to recover the test image $I$ from its coarser resolutions. 
  The resulting self-supervised network is then applied to the LR image $I$ to produce its HR output.}
}
\vspace*{-0.2cm}
\end{figure*}

\section{Image-Specific CNN}
\label{sec:images_specific_cnn}
%


Our image-specific CNN combines the predictive power and low entropy of internal image-specific information, with the generalization capabilities of Deep-Learning.
%
Given a test image $I$, with no external examples available to train on, we construct an Image-Specific CNN tailored to solve the SR task for this specific image.
We train our CNN on examples extracted from the test image itself. Such examples are obtained by downscaling the LR image $I$, to generate a lower-resolution version of itself, $I\downarrow s$ (where $s$ is the desired SR scale factor). We use a  relatively light CNN, and train it to reconstruct the test image $I$ from its lower-resolution version $I\downarrow s$ (top part of Fig.~\ref{fig:imspecific}(b)). We then apply the resulting trained CNN to the test image $I$, now using $I$ as the LR input to the network, in order to construct the desired HR output $I\uparrow s$ (bottom of Fig.~\ref{fig:imspecific}(b)). Note that the trained CNN is fully convolutional, hence can be applied to images of different sizes.

%
Since our ``training set'' consists of one instance only (the test image), we employ data augmentation on $I$ to extract more LR-HR example-pairs to train on. The augmentation is done by downscaling the test image $I$ to many smaller versions of itself ($I=I_0, I_1, I_2,...,I_n$).  These play the role of the HR supervision and are called ``HR fathers''. Each of the HR fathers is then downscaled by the desired SR scale-factor $s$ to obtain the ``LR sons'', which form the input training instances. The resulting training set consists of many image-specific LR-HR example pairs. The network can then stochastically train over these pairs.

We further enrich the training set by transforming each LR-HR pair using 4 rotations ($0^\circ, 90^\circ, 180^\circ, 270^\circ$) and their mirror reflections in the vertical and horizontal directions. This adds $\times 8$ more image-specific training examples.


For the sake of robustness, as well as to allow large SR scale factors $s$ even from very small LR images,
the SR is performed \emph{gradually}~\cite{Glasner,Timofte_2016_CVPR}.
Our algorithm is applied for several intermediate scale-factors ($s_1, s_2, ..., s_m=s$). At each intemediate scale $s_i$, we add the generated SR image HR$_i$ and its downscaled/rotated versions to our gradually growing training-set, as new HR fathers. We downscale those (as well as the previous smaller `HR examples') by the next gradual scale factor $s_{i+1}$, to generate the new LR-HR training example pairs. This is repeated until reaching the full desired resolution increase $s$. 

\subsection{Architecture \& Optimization}
\label{sec:architecture}
%
Supervised CNNs, which train on a large and diverse \emph{external} collection of LR-HR image examples, must capture
in their learned weights the large diversity of all possible LR-HR relations. As such, these networks tend to be extremely deep and very complex. In contrast, the diversity of the LR-HR relations within a single image is significantly smaller, hence can be encoded by a much smaller and simpler image-specific network.

We use a simple, fully convolutional network, with 8 hidden layers, each has 64 channels. We use ReLU activations on each layer. The network input is interpolated to the output size.
As done in previous CNN-based SR methods~\cite{DRCN,VDSR,SRCNN}, we only learn the \emph{residual} between the interpolated LR and its HR parent.
We use $L_1$ loss with ADAM optimizer~\cite{ADAM}.
We start with a learning rate of 0.001. We periodically take a linear fit of the reconstruction error and if the standard deviation is greater by a factor than the slope of the linear fit we divide the learning rate by 10. We stop when we get to a learning rate of $10^{-6}$.

To accelerate the training stage and make the \emph{runtime independent of the size} of the test image $I$, at each iteration we take a \emph{random crop of fixed size}
from a randomly-selected father-son example pair.
The crop is typically $128$$\times$$128$ (unless the sampled image-pair is smaller).
The probability of sampling a LR-HR example pair at each training iteration is set to be non-uniform
and proportional to the size of the HR-father.
The closer the size-ratio (between the HR-father and the test image $I$) is to $1$,
the higher its probability to be sampled.
This reflects the higher reliability of non-synthesized HR examples over synthesize ones.



Lastly, we use a method similar to the geometric self-ensemble proposed in~\cite{EDSR} (which generates 8 different outputs for the 8  rotations+flips of the test image $I$, and then combines them). We take the median of these 8 outputs rather than their mean. We further combine it with the back-projection technique of~\cite{IraniPeleg,Glasner}, 
so that each of the 8 output images undergoes several iterations of back-projection and finally the median image is corrected by back-projection as well. \\


\noindent
\textbf{Runtime:}
Although training is done at test time,
the average runtime per image is $54~sec$ for a single increase in SR scale-factor (on a Tesla K-80 GPU). \emph{This runtime is {independent} of the image size or the relative SR scale-factor~$s$} (this is a result of the equally sized random crops used in training; the final test runtime is negligible with respect to training iterations).
%
Nevertheless, better SR results are obtained when using a \emph{gradual} increase in resolution. For example, a gradual increase using 6 intermediate scale-factors typically improves the PSNR by $\sim$$0.2{dB}$, but increases the runtime to $\sim$$5 min$ per image.
There is therefore a tradeoff between runtime and the output quality, which is up to the user to choose.
Most of the results reported in this paper were produced using 6 intermediate scale-factors. 

For comparison, the \emph{test-time} of the leading EDSR+~\cite{EDSR} (on the same platform) is $\sim$$20~sec$ for SR$\times2$ on a $200$$\times$$200$ image. However, EDSR's run time grows \emph{quadratically} with the image size, and reaches $5 min$ for an $800 \times 800$ image. Beyond that size, our network is faster than EDSR+.

\subsection{Adapting to the Test Image}
\label{sec:adaptivity}
When the acquisition parameters of the LR  images from their HR ones are \emph{fixed} for all images (e.g., same downscaling kernel, high-quality imaging conditions), current \emph{supervised} SR methods achieve an incredible performance~\cite{NTIRE}. In practice, however, the acquisition process tends to change from image to image, since cameras/sensors differ (e.g., different lens types and PSFs), as well as the individual imaging conditions (e.g., subtle involuntary camera shake when taking the photo, poor visibility conditions, etc). This results in different downscaling kernels,
different noise characteristics, various compression artifacts, etc.
One could not practically  train for all possible image acquisition configurations/settings. 
Moreover, a single supervised CNN is unlikely to perform well for all possible types of degradations/settings. To obtain good performance, one would need many different specialized SR networks, each trained (for days or weeks) on different types of degradations/settings.

This is where the advantage of an \emph{image-specific} network comes in. Our network can be adapted to the specific degradations/settings of the test image at hand, \emph{at test time}.
Our network can receive from the user, at test time, any of the following parameters: \\
(i)~The desired downscaling kernel (when no kernel is provided, the bicubic kernel serves as a default). \\
(ii)~The desired SR scale-factor $s$. \\
(iii)~The desired number of gradual scale increases (a tradeoff between speed and quality -- the default is 6). \\
(iv)~Whether to enforce Backprojection between the LR and HR image (the default is `Yes'). \\
(v)~Whether to add `noise' to the LR sons in each LR-HR example pair extracted from the LR test image (the default is `No').

The last 2 parameters (cancelling the Backprojection and adding noise) allow to handle SR of \emph{poor-quality LR images} (whether due to sensor noise in the image, JPEG compression artifacts, etc.) We found that adding a small amount of Gaussian noise (with zero mean and a small standard-deviation of $\sim$$5$ grayscales), 
improves the performance
for a wide variety of degradations (Gaussian noise, speckle noise, JPEG artifacts, and more).
We attribute this phenomenon
to the fact that image-specific information tends to repeat across scales, whereas noise artifacts do not~\cite{Zontak:CVPR13}.
Adding a bit of synthetic noise to the LR sons (but not to their HR fathers) teaches the network to ignore uncorrelated cross-scale information (the noise), while learning to increase the resolution of correlated information (the signal details).

Indeed, our experiments show that \emph{for low-quality LR images, and for a wide variety of degradation types,
the image-specific CNN obtains significantly better SR results than SotA EDSR+~\cite{EDSR}} (see Sec.~\ref{sec:experiments_results}).
Similarly, in the case of \emph{non-ideal downscaling kernels}, the image-specific CNN obtains a significant improvement over SotA (even in the absence of any noise).
When the downscaling kernel is known (e.g., a sensor with a known PSF), it can be provided to our network. \emph{When the downscaling kernel is unknown} (which is usually the case), a {rough estimate} of the kernel can be computed directly from the test image itself (e.g., using the method of~\cite{Michaeli:ICCV13}). Such \emph{rough kernel estimations suffice to obtain $+1$$dB$ improvement over EDSR+} on non-ideal kernels (see examples in Figs.~\ref{fig:figure1} and~\ref{fig:figure2}, and empirical evaluations in Sec.~\ref{sec:experiments_results}).


Note that providing the estimated  downscaling kernel to externally-supervised SotA SR methods at test time, would be of no use. They would need to exhaustively re-train a new network on a new collection of LR-HR pairs, generated with this specific (non-parametric) downscaling kernel.

\begin{figure*}
  \centering
  \includegraphics[width=15cm]{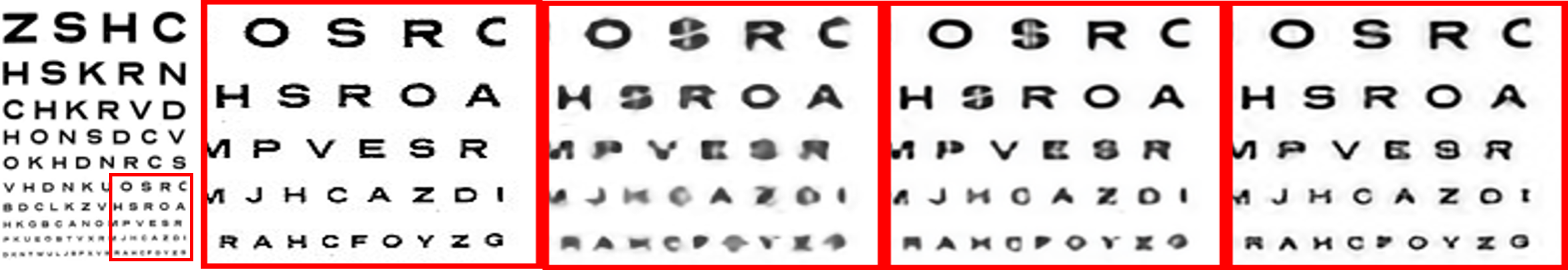}
\begin{tabular}{{p{2cm}p{2.75cm}p{2.75cm}p{2.75cm}p{2.75cm}}}
  \centering & Ground Truth & \centering VDSR \cite{VDSR} &  \centering EDSR+ \cite{EDSR} &  \centering ZSSR (ours)\tabularnewline
   \centering & (PSNR, SSIM) & \centering (20.11, 0.9136) & \centering (25.29 / 0.9627) & \centering (25.68 / 0.9546)\tabularnewline
\end{tabular}
  \caption{
\emph{In images with strong internal repetitive structures, ZSSR
tends to surpass VDSR, and sometimes also EDSR+, even
though the LR image was generated using the `ideal' supervised setting (i.e., bicubic downscaling).}
  }
  \label{fig:eyechart}
  \vspace*{0.5cm}
\end{figure*}

\begin{table*}
\begin{center}
\begin{tabular}{|c|c||c|c|c||c|c|}
\hline
\multicolumn{2}{|c||}{} & \multicolumn{3}{|c||}{Supervised} & \multicolumn{2}{|c|}{Unsupervised}\\
\hline\hline
Dataset & Scale & SRCNN \cite{SRCNN} & VDSR \cite{VDSR} & EDSR+ \cite{EDSR} & SelfExSR \cite{SelfEx} & ZSSR (ours) \\
\hline\hline
\multirow{3}{*}{Set5 }& $\times2$  & 36.66 / 0.9542 & 37.53 / 0.9587 & 38.20 / 0.9606  & 36.49 / 0.9537 & 37.37 / 0.9570\\
& $\times3$ & 32.75 / 0.9090 & 33.66 / 0.9213 & 34.76 / 0.9290 & 32.58 / 0.9093 & 33.42 / 0.9188\\
& $\times4$ & 30.48 / 0.8628 & 31.35 / 0.8838 & 32.62 / 0.8984 & 30.31 / 0.8619 & 31.13 / 0.8796\\
\hline\hline
\multirow{3}{*}{Set14}& $\times2$  & 32.42 / 0.9063 & 33.03 / 0.9124 & 34.02 / 0.9204 & 32.22 / 0.9034 & 33.00 / 0.9108\\
& $\times3$ & 29.28 / 0.8209 & 29.77 / 0.8314 & 30.66 / 0.8481 & 29.16 / 0.8196 &  29.80 / 0.8304 \\
& $\times4$ & 27.49 / 0.7503 & 28.01 / 0.7674 & 28.94 / 0.7901 & 27.40 / 0.7518 & 28.01 / 0.7651\\
\hline\hline
\multirow{3}{*}{BSD100}& $\times2$  & 31.36 / 0.8879 & 31.90 / 0.8960 & 32.37 / 0.9018 & 31.18 / 0.8855 & 31.65 / 0.8920\\
& $\times3$ & 28.41 / 0.7863 & 28.82 / 0.7976 & 29.32 / 0.8104 & 28.29 / 0.7840 &  28.67 / 0.7945 \\
& $\times4$ & 26.90 / 0.7101 &  27.29 / 0.7251 & 27.79 / 0.7437 & 26.84 / 0.7106 & 27.12 / 0.7211\\
\hline
\end{tabular}
\vspace*{-0.5cm}
\end{center}
\caption{\textbf{Comparison of SR results for the 'ideal' case (bicubic downscaling).}}
\label{table:reuslts}
\end{table*}

\begin{table*}
\begin{center}
\begin{tabular}{|c|c|c|c|c|}
\hline
VDSR \cite{VDSR} & EDSR+ \cite{EDSR} & Blind-SR~\cite{Michaeli:ICCV13} & ZSSR [estimated kernel] (ours) & ZSSR [true kernel] (ours) \\
\hline
27.7212 / 0.7635 & 27.7826 / 0.7660 & 28.420 / 0.7834 & 28.8118 / 0.8306 &29.6814 / 0.8414 \\
\hline
\end{tabular}
\vspace*{-0.5cm}
\end{center}
\caption{\small \textbf{SR in the presence of unknown downscaling kernels.}
\emph{LR images were generated from the BSD100 dataset using \emph{random} downscaling kernels (of reasonable size). SR$\times2$ was then applied to those images.  Please see text for more details.}
}
\label{table:ker_reuslts}
\end{table*}

\begin{table}
\begin{center}
\begin{tabular}{|c|c|c|}
\hline
Bicubic interpolation & EDSR+ \cite{EDSR} & ZSSR (ours) \\
\hline
27.9216 / 0.7504 & 27.5600 / 0.7135 & 28.6148 / 0.7809 \\
\hline
\end{tabular}
\vspace*{-0.5cm}
\end{center}
\caption{\small \textbf{SR in the presence of unknown image degradation.}
\emph{Each LR image from the BSD100 dataset was \emph{randomly} degraded using one of 3 types of degradations: (i)~Gaussian noise, (ii)~Speckle noise, (iii)~JPEG compression. SR$\times2$ was then applied to those images, without knowing the type of degradation.
ZSSR shows robustness to unknown degradations, whereas SotA SR methods are not. In fact, under such conditions, {bicubic interpolation outperforms current SotA SR methods.}
}
\label{table:noise_reuslts}
}
\vspace*{-0.2cm}
\end{table}

\section{Experiments \& Results}
\label{sec:experiments_results}
 Our method (ZSSR - `Zero-Shot SR') is primarily aimed at real LR images obtained with realistic (unknown and varying) acquisition setting. These usually have no HR ground truth, hence are evaluated visually (as in Fig.~\ref{fig:figure1}).
 In order to quantitatively evaluate ZSSR's performance, we ran several controlled experiments, on a variety of settings.
 %
 Interestingly, ZSSR produces \emph{competitive results} (although not SotA)  \emph{on the `ideal' benchmark datasets} for which the SotA supervised methods train and specialize (even though our CNN is small, and has not been pretrained). However, \emph{on `non-ideal' datasets, ZSSR surpasses SotA SR by a large margin}. All reported numerical results were produced using the evaluation script of~\cite{VDSR,DRCN}.

\subsection{The `Ideal' Case}
While this is not the aim of ZSSR, we tested it also on the standard SR benchmarks of `ideal' LR images. In these benchmarks, the LR images are ideally downscaled from their HR versions using MATLAB's $imresize$ command (a bicubic kernel downsampling with antialiasing).
Table~\ref{table:reuslts} shows that our image-specific ZSSR achieves competitive results against externally-supervised methods that were exhaustively trained for these conditions. In fact, ZSSR is significantly better than the older SRCNN~\cite{SRCNN}, and in some cases achieves comparable or better results than VDSR~\cite{VDSR} (which was the SotA until a year ago). Within the unsupervised-SR regime, ZSSR outperforms the leading method SelfExSR~\cite{SelfEx} by a large margin.

\begin{figure}
\includegraphics[width=8cm]{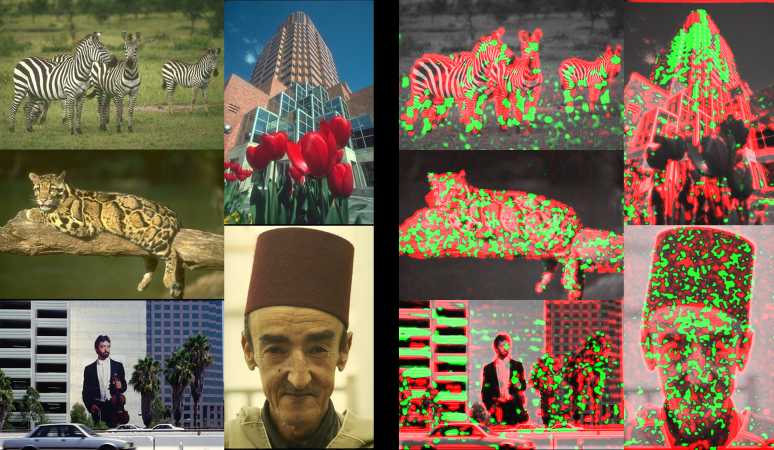}
\caption{\textbf{\textcolor{green}{Internal} vs. \textcolor{red}{External} preference.}
\emph{Green: pixels that favor Internal-SR (i.e., pixels where ZSSR obtains lower error with respect to the ground-truth HR image);
Red: pixels that favour External-SR (EDSR+). Notice that edges (which are frequent in natural images) prefer External-SR, whereas unique image structures (e.g., the tiny windows at the top of the high building) prefer Internal-SR. Such tiny patters find larger (high-res) examples of themselves elsewhere inside the same image (at a different location/scale).}}
%
\label{fig:redgreen}
\end{figure}

Moreover, in images with very strong internal repetitive structures, ZSSR tends to surpass VDSR, and sometimes also EDSR+, even though these LR images were generated using the `ideal' supervised setting. One such example is shown in Fig.~\ref{fig:eyechart}.
Although this image is not a typical natural image,  further analysis shows that the preference for \emph{internal learning} (via ZSSR) exhibited in Fig.~\ref{fig:eyechart} exists not only in `fractal-like' images, but is also found in general natural images. Several such examples are shown in Fig.~\ref{fig:redgreen}.
As can be seen, some of the pixels in the image (those marked in green) benefit more from exploiting internally learned data recurrence (ZSSR) over deeply learned external information, whereas other pixels (those marked in red) benefit more from externally learned data (EDSR+).
As expected, the internal approach (ZSSR) is mostly advantageous in image area with high recurrence of information, especially in areas where these patterns are extremely small (of extremely low resolution), like the small windows in the top of the building. Such tiny patters find larger (high-res) examples of themselves elsewhere inside the same image (at a different location/scale). This indicates that there may be potential for further SR improvement (even in the `ideal' bicubic case), by combining the power of Internal-Learning with External-Learning in a single computational framework. This remains part of our future work.

\subsection{The `Non-ideal' Case}
\label{sec:experiments_results_nonideal}
Real LR images do not tend to be ideally generated. We have experimented with non-ideal cases that result from either: (i)~non-ideal downscaling kernels (that deviate from the bicubic kernel), and (ii)~low-quality LR images (e.g., due to noise, compression artifacts, etc.)  In such non-ideal cases, the image-specific ZSSR provides significantly better results than SotA SR methods (by $1-2 dB$). These quantities experiments are described next. Fig.~\ref{fig:figure2} shows a few such visual results. Additional visual results and full images can be found in our \href{http://www.wisdom.weizmann.ac.il/~vision/zssr/}{project website}.\\

\noindent
\textbf{(A) Non-ideal downscaling kernels:} \ \
The purpose of this experiment is to test more realistic blur kernels with the ability to numerically evaluate the results. For this purpose we created a new dataset from BSD100~\cite{BSD100} by downscaling the HR images using \emph{random} (but reasonably sized) Gaussian kernels.
For each image, the covariance matrix $\bf{\Sigma}$ of its downscaling kernel was chosen to have a random angle~$\theta$ and random lengths $\lambda_1, \lambda_2$ in each axis:
\begin{align}
\lambda_1, \lambda_2 &  \backsim U[0,s^2] \nonumber \\
  \theta  & \backsim U[0,\pi] \nonumber \\
  \bf{\Lambda} & = diag(\lambda_1, \lambda_2) \nonumber \\
  \bf{U} & = \begin{bmatrix}
      \cos(\theta) & -\sin(\theta) \nonumber \\
      \sin(\theta) & \cos(\theta)
    \end{bmatrix} \nonumber \\
  \bf{\Sigma} & = \bf{U \Lambda U^t}
  \end{align}
where $s$ is the HR-LR downscaling factor. Thus, \textbf{each LR image was subsampled by a different random kernel}. 
Table~\ref{table:ker_reuslts} compares our performance against the leading externally-supervised SR methods~\cite{EDSR,VDSR}.
We also compared our performance to the unsupervised \emph{Blind-SR} method of~\cite{Michaeli:ICCV13}. We considered two cases for applying ZSSR: (i)~The more realistic scenario of \emph{unknown downscaling kernel}. For this mode we used~\cite{Michaeli:ICCV13} to evaluate the kernel directly from the test image and fed it to ZSSR. The unknown SR kernel is estimated in~\cite{Michaeli:ICCV13} by seeking a \emph{non-parametric} {downscaling kernel which maximizes the similarity of patches across scales in the LR test image}.
(ii)~We applied ZSSR with the true downscaling kernel used to create the LR image. Such a scenario is potentially useful for images obtained by sensors with  known specs.

Note that none of the externally-supervised methods are able to benefit from knowing the blur kernel of the test image (whether estimated or real), since they were trained and optimized exhaustively for a specific  kernel. Table~\ref{table:ker_reuslts} shows that ZSSR outperforms SotA methods by a very large margin: $+1db$ for unknown (estimated) kernels, and $+2db$ when provided the true kernels. Visually, the SR images generated by SotA methods are very blurry (see Fig.~\ref{fig:figure2}, and \href{http://www.wisdom.weizmann.ac.il/~vision/zssr/}{project website}) . Interestingly, the unsupervised Blind-SR method of~\cite{Michaeli:ICCV13}, which does not use any deep learning, also outperforms SotA SR methods.
This further supports the analysis and observations of~\cite{EfratLevin:ICCV13}, that (i)~an accurate donwscaling model is more important than sophisticated image priors, and (ii)~using the wrong donwscaling kernel leads to oversmoothed SR results.

A special case of a non-ideal kernel is the $\delta$ kernel, which results in aliasing. This case too, is not handled well by SotA methods (see example in Fig.~\ref{fig:figure2}). \\

\noindent
\textbf{(B) Poor-quality LR images:} \ \
In this experiment, we tested images with different types of quality degradation. To test the robustness of ZSSR in coping with \emph{unknown damage}, we chose for each image from BSD100~\cite{BSD100} a \emph{random} type of degradation out of 3 degradations: (i)~Gaussian noise [$\sigma = 0.05$], (ii)~Speckle noise [$\sigma =0.05$], (iii)~JPEG compression [quality = 45 (By MATLAB standard)].
Table~\ref{table:noise_reuslts} shows that ZSSR is robust to unknown degradation types, while these typically damage SR supervised methods to the point where \emph{bicubic interpolation outperforms current SotA SR methods!}


%

\section{Conclusion}
We introduce the concept of ``Zero-Shot'' SR, which exploits the power of Deep Learning, without relying on any external examples or prior training. This is obtained via a small \emph{image-specific CNN}, which is trained at test time on \emph{internal examples} extracted solely from the LR test image.  This yields SR of real-world images, whose acquisition process is non-ideal, unknown, and changes from image to image (i.e., image-specific settings). In such real-world `non-ideal' settings, our method substantially outperforms SotA SR methods, both qualitatively and quantitatively.
To our best knowledge, this is the first \emph{unsupervised} CNN-based SR method.
%


{\small
\bibliography{bib}{}
\bibliographystyle{ieee}
}

\end{document}